\pdfoutput=1

\documentclass[11pt]{article}

\usepackage[final]{acl}

\usepackage{times}
\usepackage{latexsym}

\usepackage[T1]{fontenc}

\usepackage[utf8]{inputenc}

\usepackage{inconsolata}

\usepackage{graphicx}
\usepackage{amsmath}
\usepackage{amsfonts}
\usepackage{bm}
\usepackage{bbm}
\usepackage{algorithm}
\usepackage{algpseudocode}
\usepackage{tcolorbox}
\usepackage{listings}
\usepackage{courier}
\usepackage{multirow}
\usepackage{booktabs}
\usepackage{microtype}
\usepackage{cleveref}

\newtheorem{theorem}{Theorem}
\newcommand{\param}{\bm{\theta}}

\title{Reward-Shifted Speculative Sampling Is An Efficient Test-Time Weak-to-Strong Aligner}

\author{
 \textbf{Bolian Li\textsuperscript{1}},
 \textbf{Yanran Wu\textsuperscript{1}},
 \textbf{Xinyu Luo\textsuperscript{1}},
 \textbf{Ruqi Zhang\textsuperscript{1}}
\\
 \textsuperscript{1}Department of Computer Science, Purdue University
\\
 \small{
   \textbf{Correspondence:} \href{mailto:li4468@purdue.edu}{li4468@purdue.edu}
 }
}

\begin{document}

\maketitle

\begin{abstract}
Aligning large language models (LLMs) with human preferences has become a critical step in their development. Recent research has increasingly focused on test-time alignment, where additional compute is allocated during inference to enhance LLM safety and reasoning capabilities. However, these test-time alignment techniques often incur substantial inference costs, limiting their practical application. We are inspired by the speculative sampling acceleration, which leverages a small draft model to efficiently predict future tokens, to address the efficiency bottleneck of test-time alignment. We introduce the reward-Shifted Speculative Sampling (SSS) algorithm, in which the draft model is aligned with human preferences, while the target model remains unchanged. We theoretically demonstrate that the distributional shift between the aligned draft model and the unaligned target model can be exploited to recover the RLHF optimal solution without actually obtaining it, by modifying the acceptance criterion and bonus token distribution. Our algorithm achieves superior gold reward scores at a significantly reduced inference cost in test-time weak-to-strong alignment experiments, thereby validating both its effectiveness and efficiency.\footnote{The implementation of this method can be found in our codebase: \url{https://github.com/lblaoke/CARDS}.}
\end{abstract}

\section{Introduction}
Large language models (LLMs) have demonstrated remarkable capabilities in tasks such as instruction-following~\citep{lou2024large}, reasoning~\citep{fei2024multimodal}, and coding~\citep{wang2024performance}. However, the practical deployment of LLMs is constrained by concerns regarding the safety and helpfulness of their outputs~\citep{bai2022training, weidinger2022taxonomy, deshpande2023toxicity}. As a result, efficiently aligning LLMs with human preferences becomes a critical challenge in LLM development. While classical reinforcement learning from human feedback (RLHF)~\citep{christiano2017deep, stiennon2020learning, ouyang2022training} has shown great potential, it often suffers from the high computational cost of RL training and is prone to instability~\citep{chen2024chatgpt, mohammadi2024creativity}.

To mitigate these issues, test-time alignment~\citep{khanov2024args, li2024cascade, qiu2024treebon} has emerged as a training-free alternative to RLHF, allocating more compute during inference to improve alignment quality. Nevertheless, most test-time alignment approaches require interaction between the LLM and an external reward model (RM) throughout decoding, resulting in either excessive LLM calls (best-of-$N$ and rejection sampling) or too many RM calls~\citep{deng2023reward, khanov2024args}. Although recent studies have recognized these inefficiencies and proposed acceleration techniques~\citep{li2024cascade, qiu2024treebon, sunfast}, these methods still rely on external RMs and require numerous calls to large models.

We are inspired by the \emph{speculative sampling} algorithm~\citep{chen2023accelerating, leviathan2023fast}, which leverages a small ``draft'' model to efficiently predict $K$ future tokens and then uses a large ``target'' model to verify them in parallel. While speculative sampling has been widely used for accelerating LLM inference, its potential for test-time alignment remains under-explored. For example, \citet{nakshatri2024constrained} applies speculative sampling to accelerate the base model inference within a reward-guided search framework. However, their approach continues to rely on an external reward model during inference, resulting in a pipeline that remains both complex and latency-prone.

We hereby propose a novel combination of test-time alignment and speculative sampling: \emph{aligning the draft model to generate high-reward tokens, while utilizing the target model for verification to ensure fluency}. This approach removes the dependence on external reward models. We introduce the reward-Shifted Speculative Sampling (SSS) algorithm, which employs an aligned draft model and an unaligned target model. Furthermore, we revise both the acceptance criterion and the residual distribution for bonus tokens, ensuring that SSS recovers the RLHF optimal solution. Extensive experiments on test-time weak-to-strong alignment tasks demonstrate that SSS achieves superior gold reward scores compared to existing baselines at a significantly reduced inference cost.

The main contributions of this paper are summarized as follows:
\begin{itemize}
    \item We redefine the objective of speculative sampling from recovering the target model distribution to recovering the RLHF optimal solution. This enables test-time alignment with a reward-shifted draft model.
    \item As a test-time alignment method, SSS eliminates the requirement for external reward models. To the best of our knowledge, this is the first approach that implements test-time alignment using only a draft and a target model.
    \item We show that SSS is an efficient algorithm for test-time weak-to-strong alignment, which enhances the alignment quality at a much lower computational cost compared to other test-time alignment methods.
\end{itemize}

\section{Preliminaries\label{sec:pre}}
\subsection{RLHF and Reward-Shifted Decoding}
The problem of LLM alignment has been modeled as a KL-constrained reward maximization process~\citep{peters2007reinforcement, korbak2022reinforcement, go2023aligning, rafailov2023direct}, in which the reward $r$ is maximized with the proximity constraint from the reference model $\pi_{\text{ref}}$:
\begin{equation}
    \max_{\bm{\theta}}\mathbb{E}_{x\in\mathcal{D}_p,y\sim\pi_{\bm{\theta}}(\cdot|x)}r(x,y) - \lambda\cdot\mathbf{KL}(\pi_{\bm{\theta}}\|\pi_{\text{ref}}).
    \label{eq:rlhf}
\end{equation}
Here, the training data only contains a prompt set $\mathcal{D}_p$, and the responses are generated by the LLM itself. The above process is the objective of reinforcement learning from human feedback (RLHF)~\citep{christiano2017deep, ouyang2022training}. We demonstrate in Section~\ref{sec:proof_rlhf} that the optimal solution of Eq.~\eqref{eq:rlhf} is:
\begin{equation}
    \pi^\star(y|x) \propto \pi_{\text{ref}}(y|x)\cdot\exp\left(\frac{1}{\beta}r(x,y)\right),
    \label{eq:rlhf_optimal}
\end{equation}
where the base model policy $\pi_{\text{ref}}$ is slightly shifted to pursue higher reward.

Built upon the above objective, another direct approach is to sample multiple responses and only keep the responses that satisfy the reward constraint, known as test-time alignment~\citep{khanov2024args, li2024cascade, qiu2024treebon}. Test-time alignment only modifies the decoding process of base models to pursue higher reward, namely a reward-shifted decoding process. There are two basic approaches to reward-shifted decoding: \romannumeral1) best-of-$N$, which generates multiple candidates in parallel and selects only the best one: $\max_{y \sim \pi_{\text{ref}}(\cdot|x)}r(x,y)$,
and \romannumeral2) rejection sampling, which continues generating proposals until a reward threshold is met: $y \sim \pi_{\text{ref}}(\cdot|x),~~~~s.t.~~r(x,y)\ge\tau_r$. These approaches are often applied to different granularities~\citep{khanov2024args, li2024cascade} and even combined together~\citep{qiu2024treebon, sunfast} for effective and efficient alignment.

\subsection{Speculative Sampling\label{sec:ss}}
Speculative sampling is an inference acceleration method for autoregressive models~\citep{chen2023accelerating}. It leverages a much smaller ``draft'' model $\pi_{\text{draft}}$ to sequentially sample candidate token sequences: $\hat{y}_{t:t+K} \sim \pi_{\text{draft}}(\cdot|x,y_{<t})$, and requires a larger and more powerful ``target'' model $\pi_{\text{ref}}$ to verify such candidates. Specifically, speculative sampling accepts a draft token with the probability:
\begin{equation}
    p_{\text{accept}}(t) = \min\left(1, \frac{\pi_{\text{ref}}(\hat{y}_t|x,y_{<t})}{\pi_{\text{draft}}(\hat{y}_t|x,y_{<t})}\right),
    \label{eq:accept_prob}
\end{equation}
where higher target model likelihoods lead to higher chances of acceptance. The complete procedure of speculative sampling is summarized in Algorithm~\ref{algo:ss}.

Speculative sampling has an acceleration trick to address the efficiency issue caused by a potentially low acceptance rate, called \emph{bonus token}. It additionally accepts one more token from the following residual distribution:
\begin{equation}
\begin{aligned}
    &\pi_{\text{bonus}}(\cdot|x,y_{<t}) \\
    &= \left(\pi_{\text{ref}}(\cdot|x,y_{<t}) - \pi_{\text{draft}}(\cdot|x,y_{<t})\right)_+,
    \label{eq:bonus}
\end{aligned}
\end{equation}
where $(f(x))_+=\frac{\max(0,f(x))}{\sum_x\max(0,f(x))}$ is the clamp normalization operator. The bonus token distribution ensures that speculative sampling recovers the target model distribution, as proved in Theorem 1 of \citet{chen2023accelerating}. This approach ensures that speculative sampling accepts at least one token per target model call, making it no slower than vanilla decoding in most cases. The procedure of speculative sampling is shown in Fig.~\ref{fig:bon_ss}.

\section{Methodology}
Most of test-time alignment methods require a reward model as the signal of human preference~\citep{khanov2024args, li2024cascade, qiu2024treebon}. The additional computation induced by such a two-model decoding process makes test-time alignment not efficient enough for time-intensive LLM serving~\citep{wang2024towards}, which calls for new frameworks to further accelerate the reward-guided decoding process. We are inspired by speculative sampling~\citep{chen2023accelerating}, which accelerates autoregressive decoding via a draft-then-verify procedure. We propose to shift the small draft model to align with human preferences (computationally cheap) and design a new speculative sampling algorithm to simulate the distribution of an aligned target model without actually obtaining it. We also prove that the proposed algorithm recovers the RLHF optimal solution.

\subsection{Shifting Draft Models to Align with Human Preferences}
The first step of the proposed framework is obtaining a well-aligned draft model to reflect human preference. Following the settings of \citet{tao2024your}, we start from a SFT checkpoint $\pi_{\text{draft}}^{\text{SFT}}$ fine-tuned on the chosen responses of preference data, and align this model via direct preference optimization (DPO)~\citep{rafailov2023direct}. We assume that the aligned draft model $\pi_{\text{draft}}^r$ follows the RLHF optimal solution for $\pi_{\text{draft}}^{\text{SFT}}$:
\begin{equation}
    \pi_{\text{draft}}^r(y|x) \approx \pi_{\text{draft}}^{\text{SFT}}(y|x)\cdot\exp\left(\frac{1}{\beta}r(x,y)\right).
\end{equation}

\begin{figure}[t]
    \centering
    \includegraphics[width=\linewidth]{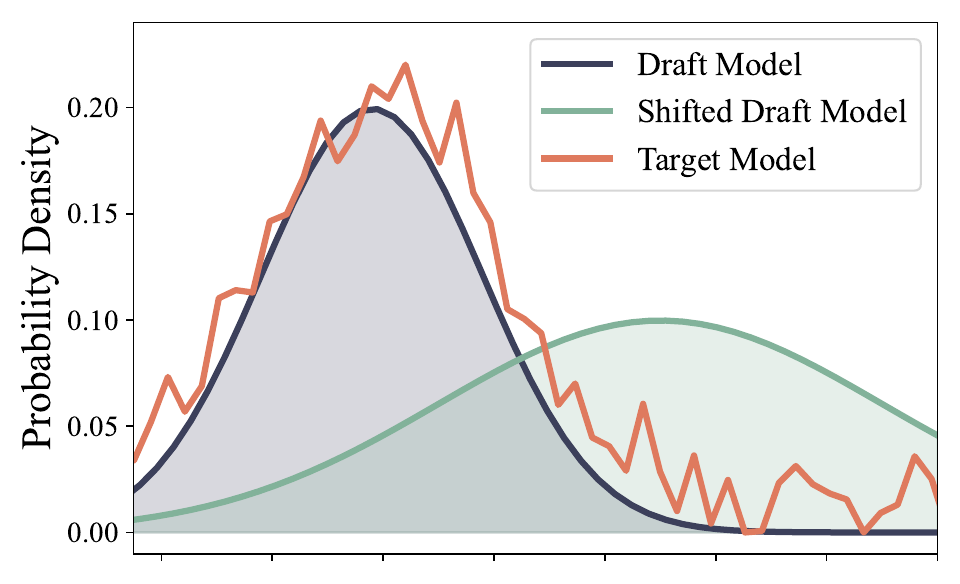}
    \caption{Shifted draft model reflects human preferences, but presents a gap from unaligned target model.}
    \label{fig:dist_shift}
\end{figure}

However, shifting the draft model will enlarge its gap from the target model, and consequently lowers the acceptance rate~\citep{hongtraining}. As visualized in Fig.~\ref{fig:dist_shift}, aligning draft models to human preference is at the cost of a significant distributional shift. Directly applying the shifted draft model $\pi_{\text{draft}}^r$ to standard speculative sampling (Algorithm~\ref{algo:ss}) would result in severe efficiency issues. This is because standard speculative sampling only recovers the unaligned target model's distribution, leading to a low acceptance rate due to the distribution shift between the draft and target models, as shown in Table~\ref{tab:accept_drop}.
 This drawback motivates us to propose a new speculative sampling algorithm to raise the acceptance rate and ensure that generated responses recover the aligned target model's distribution without actually obtaining it.

\begin{table}[t]\small\setlength\tabcolsep{5.2pt}\renewcommand{\arraystretch}{0.8}
\centering
\caption{Standard speculative sampling with a shifted draft model suffers from low acceptance rate.}
\label{tab:accept_drop}
\begin{tabular}{lll}
\toprule
\multicolumn{1}{c}{\textbf{Target Model}} & \multicolumn{1}{c}{\textbf{Draft Model}} & \multicolumn{1}{c}{\textbf{Acceptance Rate}} \\ \midrule
\multirow{2}{*}{\texttt{OPT-6.7B}}        & \texttt{OPT-125M}                        & 0.33                                         \\
                                          & \texttt{OPT-125M} aligned                & 0.08 (\textcolor{red}{$\downarrow$ 76\%})    \\ \midrule
\multirow{2}{*}{\texttt{OPT-13B}}         & \texttt{OPT-350M}                        & 0.42                                         \\
                                          & \texttt{OPT-350M} aligned                & 0.13 (\textcolor{red}{$\downarrow$ 69\%})    \\ \bottomrule
\end{tabular}
\end{table}

\subsection{Speculative Sampling under Draft-Target Distributional Shift}
The aforementioned distributional shift caused by aligning draft models to human preferences is a critical problem in reward-shifted speculative sampling. To resolve this, we propose a new speculative sampling algorithm that utilizes such a distributional shift to recover the optimal solution of RLHF, which is typically obtainable by training the target model on preference data~\citep{schulman2017proximal, rafailov2023direct, shao2024deepseekmath}. We eliminate the need for training target models and significantly reduce the decoding cost compared with previous test-time alignment methods~\citep{khanov2024args, li2024cascade, qiu2024treebon}.

As discussed in Section~\ref{sec:ss}, the bonus token sampled from the residual distribution $\pi_{\text{bonus}}$ (Eq.~\eqref{eq:bonus}) guarantees that standard speculative sampling recovers the distribution of target model $\pi_{\text{ref}}$~\citep{chen2023accelerating}. Intuitively, it compensates for the influence of draft model distribution $\pi_{\text{draft}}$ when draft tokens are rejected. In our proposed new speculative sampling algorithm, the acceptance probability is modified as:
\begin{equation}
    p_{\text{accept}}(t) = \min\left(1, \frac{\pi_{\text{ref}}(\hat{y}_t|x,y_{<t})}{\pi_{\text{draft}}^{\text{SFT}}(\hat{y}_t|x,y_{<t})}\right),
    \label{eq:accept_prob_new}
\end{equation}
and the residual distribution for bonus tokens also has a new form:
\begin{equation}
\begin{aligned}
    &\pi_{\text{bonus}}^r(\cdot|x,y_{<t+K'}) \\
    &=\left(\pi_{\text{draft}}^r(\cdot|x,y_{<t+K'}) \left(\frac{\pi_{\text{ref}}(\cdot|x,y_{<t+K'})}{\pi_{\text{draft}}^{\text{SFT}}(\cdot|x,y_{<t+K'})} - 1\right)\right)_+,
    \label{eq:r_bonus}
\end{aligned}
\end{equation}
where $K'$ is the number of actually accepted draft tokens. The entire speculative sampling process is detailed in Algorithm~\ref{algo:sss}. Compared to the standard version (Section~\ref{sec:ss}), the condition and reactions for rejecting draft tokens are modified, and no additional token is sampled from the target model once all draft tokens are accepted, since our objective is not the target model distribution.

Our new speculative sampling handles inference acceleration and preference alignment simultaneously. It leverages a shifted draft model $\pi_{\text{draft}}^r$ to generate human-preferred draft tokens, and uses a unaligned target model $\pi_{\text{ref}}$ to ensure the fluency of responses. We also guarantee that \emph{the proposed algorithm recovers the RLHF optimal solution} (Eq.~\eqref{eq:rlhf_optimal}), as discussed in Theorem~\ref{theorem} and proved in Section~\ref{sec:proof_theorem}.

\begin{theorem}[SSS recovers RLHF optimal solution]
\label{theorem}
Shifted speculative sampling (SSS) as demonstrated in Algorithm~\ref{algo:sss} recovers the RLHF optimal solution in Eq.~\eqref{eq:rlhf_optimal}. Specifically, assume a well-aligned draft model $\pi_{\text{draft}}^r(y|x) = \pi_{\text{draft}}^{\text{SFT}}(y|x)\cdot\exp\left(\frac{1}{\beta}r(x,y)\right)$, the probability that SSS generates response $y$ given prompt $x$ is:
\[
\mathbf{P}(Y=y|x) \equiv \pi^\star(y|x).
\]
\end{theorem}

\begin{algorithm}[t]
\caption{Shifted Speculative Sampling}
\label{algo:sss}
\begin{algorithmic}

\Require autoregressive target model $\pi_{\text{ref}}$ and autoregressive $r$-shifted draft model $\pi_{\text{draft}}^r$;
\Require initial prompt $x$ and lookahead $K>0$.

\State $y \gets \emptyset$
\State $t \gets 0$

\While{$t < \texttt{max\_length}$}
    \For{k = 1 : K}
        \State $\hat{y}_{t+k} \sim \pi_{\text{draft}}^r(\cdot|x,y_{<t},\hat{y}_{t:t+k-1})$
    \EndFor
    \State $\pi_{\text{ref}}(\hat{y}_{t:t+K}|x,y_{<t})$ \Comment{Target Likelihoods}

    \For{k = 1 : K}
        \State $\epsilon \sim \mathcal{U}[0,1]$
        \If{$\epsilon < p_{\text{accept}}(t)$ in Eq.~\eqref{eq:accept_prob_new}}
            \State $y_{t+k} \gets \hat{y}_{t+k}$ \Comment{Accept}
        \Else
            \State $y_{t+k} \gets \pi_{\text{bonus}}^r(\cdot|x,y_{<t+k})$ in Eq.~\eqref{eq:r_bonus}
            \State $t \gets t+k$ and break
        \EndIf
    \EndFor

    \If{all draft tokens are accepted}
        \State $t \gets t+K$
    \EndIf
\EndWhile

\end{algorithmic}
\end{algorithm}

\section{Experiments\label{sec:exp}}
In this section, we present the experimental settings and empirical results. We first discuss the configurations of experiments in Section~\ref{sec:exp_setting}, then show the main empirical results in Section~\ref{sec:exp_result}, and finally have ablation studies on the draft model selection and alternative algorithm options in Section~\ref{sec:exp_ablation}.

\subsection{Experimental Settings\label{sec:exp_setting}}
All experiments in this paper are based on the HH-RLHF dataset~\citep{ganguli2022red}, which contains paired conversational data on helpfulness and harmlessness to reflect human preference. To ensure that draft models and target models share the same vocabulary and have similar distributions, we choose the 3 model pairs as shown in Table~\ref{tab:hyper_model}. For the gold reward in response evaluation, we choose a large reward model trained on HH-RLHF with \texttt{Llama-7B} backbone~\citep{llama7brm2024}.

\begin{table}[ht]\small\renewcommand{\arraystretch}{0.8}
\centering
\caption{Base model choices for draft and target models.}
\label{tab:hyper_model}
\begin{tabular}{ll}
\toprule
\multicolumn{1}{c}{\textbf{Draft Model}} & \multicolumn{1}{c}{\textbf{Target Model}} \\ \midrule
\texttt{Qwama-0.5B}                      & \texttt{Llama-3-8B}                       \\
\texttt{OPT-125M}                        & \texttt{OPT-6.7B}                         \\
\texttt{OPT-350M}                        & \texttt{OPT-13B}                          \\ \bottomrule
\end{tabular}
\end{table}

For post-training draft models, we use the TRL library~\citep{vonwerra2022trl} for supervised fine-tuning (SFT) and direct preference optimization (DPO)~\citep{rafailov2023direct} with the hyper-parameters in Table~\ref{tab:hyper_draft_training}. These hyper-parameter choices are based on the settings of \citet{tao2024your} and adjusted for better performance by grid search.

\begin{table}[ht]\small\setlength\tabcolsep{4.8pt}\renewcommand{\arraystretch}{0.8}
\centering
\caption{Hyper-parameters for fine-tuning and post-training draft models on preference data.}
\label{tab:hyper_draft_training}
\begin{tabular}{lc|ccc}
\toprule
\multicolumn{1}{c}{\textbf{Base Model}} & \textbf{Pipeline} & \textbf{\begin{tabular}[c]{@{}c@{}}Learning\\ Rate\end{tabular}} & \textbf{\begin{tabular}[c]{@{}c@{}}Batch\\ Size\end{tabular}} & \textbf{Epochs} \\ \midrule
\multirow{2}{*}{\texttt{Qwama-0.5B}}    & SFT               & 2e-4                                                             & 32                                                            & 1.00            \\
                                        & DPO               & 1e-5                                                             & 16                                                            & 0.57            \\ \midrule
\multirow{2}{*}{\texttt{OPT-125M}}      & SFT               & 2e-6                                                             & 32                                                            & 1.00            \\
                                        & DPO               & 5e-5                                                             & 16                                                            & 1.00            \\ \midrule
\multirow{2}{*}{\texttt{OPT-350M}}      & SFT               & 2e-6                                                             & 32                                                            & 0.57            \\
                                        & DPO               & 5e-5                                                             & 16                                                            & 1.00            \\ \bottomrule
\end{tabular}
\end{table}

For inference experiments, we use the Transformers library~\citep{wolf-etal-2020-transformers} with a temperature of 0.8 and a maximum sequence length of 128 tokens. These settings are standard in previous works~\citep{chen2023accelerating, khanov2024args, li2024cascade}.

Additionally, due to the small size of draft models, all experiments can be conducted on one NVIDIA L40S GPU. Time measurements are obtained from the \texttt{time.time()} API in Python.

\subsection{Alignment Quality and Inference Efficiency\label{sec:exp_result}}
\Cref{tab:reward_comparison} demonstrates the performance advantages of SSS across different model pairs on HH-RLHF for a single run. We compare against a range of test-time alignment baselines, including Best‑of‑$N$ (BoN), TreeBoN~\citep{qiu2024treebon}, and CARDS~\citep{li2024cascade}. We keep the reward model sizes used in these baselines the same as draft models for fair comparison. The baseline choices represent a spectrum of test-time alignment strategies balancing alignment quality and computational efficiency. We also include comparisons with a standard speculative sampling (SS) approach.

Our method, SSS, consistently achieves the best trade-off between gold reward and efficiency. For example, on \texttt{OPT-6.7B}, SSS achieves the highest gold reward (3.88) while requiring only 115 LLM calls and 7.8 seconds per response, which has a 2.9$\times$ speedup over BoN. On \texttt{Llama-3-8B}, SSS reduces inference time by over 5$\times$ compared to BoN, while maintaining competitive gold reward. Compared to CARDS and TreeBoN, SSS offers significantly lower inference cost and latency, with high alignment quality. Among all baselines, our method achieves the highest gold reward on \texttt{OPT-13B}, reaching 4.06 while also delivering a 5.1$\times$ speedup over BoN, highlighting its strong alignment quality and inference efficiency.

\begin{table}[ht]\small\setlength{\tabcolsep}{4.6pt}\renewcommand{\arraystretch}{0.8}
\centering
\caption{\textbf{Comparison of test-time weak-to-strong alignment methods in terms of gold reward, number of LLM calls, and inference time.} ``Gold R'' refers to the average score assigned by the gold reward model, ``\# Calls'' indicates the number of forward passes through the target model $\pi_{\text{ref}}$, and ``Time'' is the actual wall-clock inference time per response. ``Speedup'' is computed relative to the BoN baseline for each method.}
\begin{tabular}{c|l|cccc}
\toprule
& \textbf{Method} & \textbf{Gold R} & \textbf{\# Calls} & \textbf{Time (s)} & \textbf{Speedup}\\
\midrule

\multirow{6}{*}{\rotatebox{90}{\texttt{Llama-3-8B}}} 
& Vanilla     & 5.63 & 128.0 & 6.5 & \textendash\\
& BoN-10      & 6.37 & 1280 & 58.0 & 1.0$\times$\\
& TreeBoN     & \textbf{6.44} & 841.4 & 48.1 & 1.2$\times$\\
& CARDS       & 6.41 & 790.7 & 45.7 & 1.3$\times$\\
& Vanilla SD  & 5.74 & 58.6    & 7.6 &\textbf{7.6}$\times$\\
& \textbf{SSS (our)}  & 6.14 & 86.2 & 10.7  &5.4$\times$\\
\midrule

\multirow{6}{*}{\rotatebox{90}{\texttt{OPT-6.7B}}} 
& Vanilla         & 3.21 & 128.0    &5.5 & \textendash\\
& BoN-5 & 3.28 & 640.0    & 22.7 & 1.0$\times$\\
& TreeBoN & 3.27 & 801.6    & 25.6 &0.9$\times$\\
& CARDS & 2.93 & 414.1    & 12.9 &1.8$\times$\\
& Vanilla SD & 2.00 & 41.7    & 3.1 &\textbf{7.4}$\times$\\
& \textbf{SSS (our)}   & \textbf{3.88} & 115    & 7.8 &2.9$\times$\\
\midrule

\multirow{6}{*}{\rotatebox{90}{\texttt{OPT-13B}}} 
& Vanilla         & 3.13 & 128.0    &9.5 & \textendash\\
& BoN-5 & 3.49 & 640.0    & 69.4 & 1.0$\times$\\
& TreeBoN & 3.61 & 968.6    & 69.4 &1.0$\times$\\
& CARDS & 3.35 & 741.9    & 50.1 &1.4$\times$\\
& Vanilla SD & 3.85 & 108.5    & 14.9 &4.7$\times$\\
& \textbf{SSS (our)}   & \textbf{4.06} & 112.5    & 13.6 &\textbf{5.1}$\times$\\
\bottomrule
\end{tabular}
\label{tab:reward_comparison}
\end{table}

\section{Conclusion and Limitations}
This paper introduces a novel speculative sampling algorithm designed for efficient test-time alignment. Our approach involves shifting the draft model to align with human preferences, thereby intentionally creating a distributional shift between the draft and target models. This shift is leveraged to simulate the distribution of a well-aligned target model (i.e., the RLHF optimal solution). We modify the standard speculative sampling algorithm by introducing a new acceptance criterion and a new bonus token distribution, ensuring that our algorithm recovers the RLHF optimal solution. Compared to existing test-time weak-to-strong alignment methods, our algorithm achieves superior alignment quality (measured by gold reward) while substantially reducing inference costs.

Despite these promising results, the effectiveness of our algorithm depends on the assumption that the shifted draft model is well-aligned: $\pi_{\text{draft}}^r(y|x) \approx \pi_{\text{draft}}^r(y|x)\cdot\exp\left(\frac{1}{\beta}r(x,y)\right)$. This assumption is sensitive to the post-training process of the draft models. Accurately verifying this assumption is also challenging due to the unknown reward function, making the empirical performance of our algorithm contingent on the tuning of hyper-parameters during draft model post-training. Furthermore, employing a small draft model to capture human preferences may lead to generalization issues: a draft model aligned for one task may not perform well on others. We plan to address these limitations in future works.


\section*{Statement of Reproducibility}
The code used for inference experiments is included in the supplementary material. It will be publicly available upon acceptance.

\section*{Statement of AI Assistant Usage}
The construction of the codebase partially relied on AI assistant for debugging, and the writing of this paper was polished by AI assistant.

\small
\bibliography{custom}

\begin{thebibliography}{40}
\providecommand{\natexlab}[1]{#1}

\bibitem[{argsearch(2024)}]{llama7brm2024}
argsearch. 2024.
\newblock Llama 7b reward model (float32).
\newblock \url{https://huggingface.co/argsearch/llama-7b-rm-float32}.
\newblock Accessed: 2025-05-01.

\bibitem[{Bai et~al.(2022)Bai, Jones, Ndousse, Askell, Chen, DasSarma, Drain, Fort, Ganguli, Henighan et~al.}]{bai2022training}
Yuntao Bai, Andy Jones, Kamal Ndousse, Amanda Askell, Anna Chen, Nova DasSarma, Dawn Drain, Stanislav Fort, Deep Ganguli, Tom Henighan, and 1 others. 2022.
\newblock Training a helpful and harmless assistant with reinforcement learning from human feedback.
\newblock \emph{arXiv preprint arXiv:2204.05862}.

\bibitem[{Cai et~al.(2024)Cai, Li, Geng, Peng, Lee, Chen, and Dao}]{cai2024medusa}
Tianle Cai, Yuhong Li, Zhengyang Geng, Hongwu Peng, Jason~D Lee, Deming Chen, and Tri Dao. 2024.
\newblock Medusa: Simple llm inference acceleration framework with multiple decoding heads.
\newblock In \emph{International Conference on Machine Learning}, pages 5209--5235. PMLR.

\bibitem[{Chen et~al.(2023)Chen, Borgeaud, Irving, Lespiau, Sifre, and Jumper}]{chen2023accelerating}
Charlie Chen, Sebastian Borgeaud, Geoffrey Irving, Jean-Baptiste Lespiau, Laurent Sifre, and John Jumper. 2023.
\newblock Accelerating large language model decoding with speculative sampling.
\newblock \emph{arXiv preprint arXiv:2302.01318}.

\bibitem[{Chen et~al.(2024)Chen, Zaharia, and Zou}]{chen2024chatgpt}
Lingjiao Chen, Matei Zaharia, and James Zou. 2024.
\newblock How is chatgpt’s behavior changing over time?
\newblock \emph{Harvard Data Science Review}, 6(2).

\bibitem[{Christiano et~al.(2017)Christiano, Leike, Brown, Martic, Legg, and Amodei}]{christiano2017deep}
Paul~F Christiano, Jan Leike, Tom Brown, Miljan Martic, Shane Legg, and Dario Amodei. 2017.
\newblock Deep reinforcement learning from human preferences.
\newblock \emph{Advances in neural information processing systems}, 30.

\bibitem[{Deng and Raffel(2023)}]{deng2023reward}
Haikang Deng and Colin Raffel. 2023.
\newblock Reward-augmented decoding: Efficient controlled text generation with a unidirectional reward model.
\newblock In \emph{Conference on Empirical Methods in Natural Language Processing}.

\bibitem[{Deshpande et~al.(2023)Deshpande, Murahari, Rajpurohit, Kalyan, and Narasimhan}]{deshpande2023toxicity}
Ameet Deshpande, Vishvak Murahari, Tanmay Rajpurohit, Ashwin Kalyan, and Karthik~R Narasimhan. 2023.
\newblock Toxicity in chatgpt: Analyzing persona-assigned language models.
\newblock In \emph{Empirical Methods in Natural Language Processing}.

\bibitem[{Elhoushi et~al.(2024)Elhoushi, Shrivastava, Liskovich, Hosmer, Wasti, Lai, Mahmoud, Acun, Agarwal, Roman et~al.}]{elhoushi2024layerskip}
Mostafa Elhoushi, Akshat Shrivastava, Diana Liskovich, Basil Hosmer, Bram Wasti, Liangzhen Lai, Anas Mahmoud, Bilge Acun, Saurabh Agarwal, Ahmed Roman, and 1 others. 2024.
\newblock Layerskip: Enabling early exit inference and self-speculative decoding.
\newblock In \emph{Proceedings of the 62nd Annual Meeting of the Association for Computational Linguistics (Volume 1: Long Papers)}, pages 12622--12642.

\bibitem[{Fei et~al.(2024)Fei, Yao, Zhang, Liu, Zhang, and Chua}]{fei2024multimodal}
Hao Fei, Yuan Yao, Zhuosheng Zhang, Fuxiao Liu, Ao~Zhang, and Tat-Seng Chua. 2024.
\newblock From multimodal llm to human-level ai: Modality, instruction, reasoning, efficiency and beyond.
\newblock In \emph{Proceedings of the 2024 Joint International Conference on Computational Linguistics, Language Resources and Evaluation (LREC-COLING 2024): Tutorial Summaries}, pages 1--8.

\bibitem[{Fu et~al.(2025)Fu, Bailis, Stoica, and Zhang}]{fu2024lookahead}
Yichao Fu, Peter Bailis, Ion Stoica, and Hao Zhang. 2025.
\newblock Break the sequential dependency of llm inference using lookahead decoding.
\newblock In \emph{Forty-first International Conference on Machine Learning}.

\bibitem[{Ganguli et~al.(2022)Ganguli, Lovitt, Kernion, Askell, Bai, Kadavath, Mann, Perez, Schiefer, Ndousse et~al.}]{ganguli2022red}
Deep Ganguli, Liane Lovitt, Jackson Kernion, Amanda Askell, Yuntao Bai, Saurav Kadavath, Ben Mann, Ethan Perez, Nicholas Schiefer, Kamal Ndousse, and 1 others. 2022.
\newblock Red teaming language models to reduce harms: Methods, scaling behaviors, and lessons learned.
\newblock \emph{arXiv preprint arXiv:2209.07858}.

\bibitem[{Go et~al.(2023)Go, Korbak, Kruszewski, Rozen, Ryu, and Dymetman}]{go2023aligning}
Dongyoung Go, Tomasz Korbak, Germ{\`a}n Kruszewski, Jos Rozen, Nahyeon Ryu, and Marc Dymetman. 2023.
\newblock Aligning language models with preferences through $ f $-divergence minimization.
\newblock In \emph{International Conference on Machine Learning}, pages 11546--11583. PMLR.

\bibitem[{Hong et~al.(2025)Hong, Raju, Li, Li, Thakker, Ravichandran, Jain, and Hu}]{hongtraining}
Fenglu Hong, Ravi~Shanker Raju, Jonathan~Lingjie Li, Bo~Li, Urmish Thakker, Avinash Ravichandran, Swayambhoo Jain, and Changran Hu. 2025.
\newblock Training domain draft models for speculative decoding: Best practices and insights.
\newblock In \emph{First Workshop on Scalable Optimization for Efficient and Adaptive Foundation Models}.

\bibitem[{Khanov et~al.(2024)Khanov, Burapacheep, and Li}]{khanov2024args}
Maxim Khanov, Jirayu Burapacheep, and Yixuan Li. 2024.
\newblock Args: Alignment as reward-guided search.
\newblock In \emph{The Twelfth International Conference on Learning Representations}.

\bibitem[{Korbak et~al.(2022)Korbak, Elsahar, Kruszewski, and Dymetman}]{korbak2022reinforcement}
Tomasz Korbak, Hady Elsahar, Germ{\'a}n Kruszewski, and Marc Dymetman. 2022.
\newblock On reinforcement learning and distribution matching for fine-tuning language models with no catastrophic forgetting.
\newblock \emph{Advances in Neural Information Processing Systems}, 35:16203--16220.

\bibitem[{Leviathan et~al.(2023)Leviathan, Kalman, and Matias}]{leviathan2023fast}
Yaniv Leviathan, Matan Kalman, and Yossi Matias. 2023.
\newblock Fast inference from transformers via speculative decoding.
\newblock In \emph{International Conference on Machine Learning}, pages 19274--19286. PMLR.

\bibitem[{Li et~al.(2025)Li, Wang, Lochab, Grama, and Zhang}]{li2024cascade}
Bolian Li, Yifan Wang, Anamika Lochab, Ananth Grama, and Ruqi Zhang. 2025.
\newblock Cascade reward sampling for efficient decoding-time alignment.
\newblock \emph{Conference of Language Modeling}.

\bibitem[{Li et~al.(2024)Li, Wei, Zhang, and Zhang}]{li2024eagle}
Yuhui Li, Fangyun Wei, Chao Zhang, and Hongyang Zhang. 2024.
\newblock Eagle: Speculative sampling requires rethinking feature uncertainty.
\newblock In \emph{International Conference on Machine Learning}, pages 28935--28948. PMLR.

\bibitem[{Liao et~al.(2025)Liao, Xu, Dong, Li, Monz, Savarese, Sahoo, and Xiong}]{liao2025reward}
Baohao Liao, Yuhui Xu, Hanze Dong, Junnan Li, Christof Monz, Silvio Savarese, Doyen Sahoo, and Caiming Xiong. 2025.
\newblock Reward-guided speculative decoding for efficient llm reasoning.
\newblock In \emph{Forty-second International Conference on Machine Learning}.

\bibitem[{Liu et~al.(2024)Liu, Hu, Bailis, Cheung, Deng, Stoica, and Zhang}]{liu2024online}
Xiaoxuan Liu, Lanxiang Hu, Peter Bailis, Alvin Cheung, Zhijie Deng, Ion Stoica, and Hao Zhang. 2024.
\newblock Online speculative decoding.
\newblock In \emph{Proceedings of the 41st International Conference on Machine Learning}, pages 31131--31146.

\bibitem[{Lou et~al.(2024)Lou, Zhang, and Yin}]{lou2024large}
Renze Lou, Kai Zhang, and Wenpeng Yin. 2024.
\newblock Large language model instruction following: A survey of progresses and challenges.
\newblock \emph{Computational Linguistics}, 50(3):1053--1095.

\bibitem[{Miao et~al.(2024)Miao, Oliaro, Zhang, Cheng, Wang, Zhang, Wong, Zhu, Yang, Shi et~al.}]{miao2024specinfer}
Xupeng Miao, Gabriele Oliaro, Zhihao Zhang, Xinhao Cheng, Zeyu Wang, Zhengxin Zhang, Rae Ying~Yee Wong, Alan Zhu, Lijie Yang, Xiaoxiang Shi, and 1 others. 2024.
\newblock Specinfer: Accelerating large language model serving with tree-based speculative inference and verification.
\newblock In \emph{Proceedings of the 29th ACM International Conference on Architectural Support for Programming Languages and Operating Systems, Volume 3}, pages 932--949.

\bibitem[{Mohammadi(2024)}]{mohammadi2024creativity}
Behnam Mohammadi. 2024.
\newblock Creativity has left the chat: The price of debiasing language models.
\newblock \emph{Available at SSRN 4858364}.

\bibitem[{Nakshatri et~al.(2025)Nakshatri, Roy, Das, Chaidaroon, Boytsov, and Gangadharaiah}]{nakshatri2024constrained}
Nishanth~Sridhar Nakshatri, Shamik Roy, Rajarshi Das, Suthee Chaidaroon, Leonid Boytsov, and Rashmi Gangadharaiah. 2025.
\newblock Constrained decoding with speculative lookaheads.
\newblock In \emph{Proceedings of the 2025 Conference of the Nations of the Americas Chapter of the Association for Computational Linguistics: Human Language Technologies (Volume 1: Long Papers)}, pages 4681--4700.

\bibitem[{Ouyang et~al.(2022)Ouyang, Wu, Jiang, Almeida, Wainwright, Mishkin, Zhang, Agarwal, Slama, Ray et~al.}]{ouyang2022training}
Long Ouyang, Jeffrey Wu, Xu~Jiang, Diogo Almeida, Carroll Wainwright, Pamela Mishkin, Chong Zhang, Sandhini Agarwal, Katarina Slama, Alex Ray, and 1 others. 2022.
\newblock Training language models to follow instructions with human feedback.
\newblock \emph{Advances in neural information processing systems}, 35:27730--27744.

\bibitem[{Peters and Schaal(2007)}]{peters2007reinforcement}
Jan Peters and Stefan Schaal. 2007.
\newblock Reinforcement learning by reward-weighted regression for operational space control.
\newblock In \emph{Proceedings of the 24th international conference on Machine learning}, pages 745--750.

\bibitem[{Qiu et~al.(2024)Qiu, Lu, Zeng, Guo, Geng, Wang, Huang, Wu, and Wang}]{qiu2024treebon}
Jiahao Qiu, Yifu Lu, Yifan Zeng, Jiacheng Guo, Jiayi Geng, Huazheng Wang, Kaixuan Huang, Yue Wu, and Mengdi Wang. 2024.
\newblock Treebon: Enhancing inference-time alignment with speculative tree-search and best-of-n sampling.
\newblock \emph{arXiv preprint arXiv:2410.16033}.

\bibitem[{Rafailov et~al.(2023)Rafailov, Sharma, Mitchell, Manning, Ermon, and Finn}]{rafailov2023direct}
Rafael Rafailov, Archit Sharma, Eric Mitchell, Christopher~D Manning, Stefano Ermon, and Chelsea Finn. 2023.
\newblock Direct preference optimization: Your language model is secretly a reward model.
\newblock \emph{Advances in Neural Information Processing Systems}, 36:53728--53741.

\bibitem[{Schulman et~al.(2017)Schulman, Wolski, Dhariwal, Radford, and Klimov}]{schulman2017proximal}
John Schulman, Filip Wolski, Prafulla Dhariwal, Alec Radford, and Oleg Klimov. 2017.
\newblock Proximal policy optimization algorithms.
\newblock \emph{arXiv preprint arXiv:1707.06347}.

\bibitem[{Shao et~al.(2024)Shao, Wang, Zhu, Xu, Song, Bi, Zhang, Zhang, Li, Wu et~al.}]{shao2024deepseekmath}
Zhihong Shao, Peiyi Wang, Qihao Zhu, Runxin Xu, Junxiao Song, Xiao Bi, Haowei Zhang, Mingchuan Zhang, YK~Li, Y~Wu, and 1 others. 2024.
\newblock Deepseekmath: Pushing the limits of mathematical reasoning in open language models.
\newblock \emph{arXiv preprint arXiv:2402.03300}.

\bibitem[{Stiennon et~al.(2020)Stiennon, Ouyang, Wu, Ziegler, Lowe, Voss, Radford, Amodei, and Christiano}]{stiennon2020learning}
Nisan Stiennon, Long Ouyang, Jeffrey Wu, Daniel Ziegler, Ryan Lowe, Chelsea Voss, Alec Radford, Dario Amodei, and Paul~F Christiano. 2020.
\newblock Learning to summarize with human feedback.
\newblock \emph{Advances in neural information processing systems}, 33:3008--3021.

\bibitem[{Sun et~al.(2024)Sun, Haider, Zhang, Yang, Qiu, Yin, Wang, Bartlett, and Zanette}]{sunfast}
Hanshi Sun, Momin Haider, Ruiqi Zhang, Huitao Yang, Jiahao Qiu, Ming Yin, Mengdi Wang, Peter Bartlett, and Andrea Zanette. 2024.
\newblock Fast best-of-n decoding via speculative rejection.
\newblock In \emph{The Thirty-eighth Annual Conference on Neural Information Processing Systems}.

\bibitem[{Tao and Li(2025)}]{tao2024your}
Leitian Tao and Yixuan Li. 2025.
\newblock Your weak llm is secretly a strong teacher for alignment.
\newblock In \emph{The Thirteenth International Conference on Learning Representations}.

\bibitem[{von Werra et~al.(2020)von Werra, Belkada, Tunstall, Beeching, Thrush, Lambert, Huang, Rasul, and Gallouédec}]{vonwerra2022trl}
Leandro von Werra, Younes Belkada, Lewis Tunstall, Edward Beeching, Tristan Thrush, Nathan Lambert, Shengyi Huang, Kashif Rasul, and Quentin Gallouédec. 2020.
\newblock Trl: Transformer reinforcement learning.
\newblock \url{https://github.com/huggingface/trl}.

\bibitem[{Wang et~al.(2024{\natexlab{a}})Wang, Shi, Du, Tao, Shen, Zheng, and Qiu}]{wang2024performance}
Lun Wang, Chuanqi Shi, Shaoshuai Du, Yiyi Tao, Yixian Shen, Hang Zheng, and Xinyu Qiu. 2024{\natexlab{a}}.
\newblock Performance review on llm for solving leetcode problems.
\newblock In \emph{2024 4th International Symposium on Artificial Intelligence and Intelligent Manufacturing (AIIM)}, pages 1050--1054. IEEE.

\bibitem[{Wang et~al.(2024{\natexlab{b}})Wang, Chen, Li, Tang, Guo, Wang, Wang, Zhou, and Chu}]{wang2024towards}
Yuxin Wang, Yuhan Chen, Zeyu Li, Zhenheng Tang, Rui Guo, Xin Wang, Qiang Wang, Amelie~Chi Zhou, and Xiaowen Chu. 2024{\natexlab{b}}.
\newblock Towards efficient and reliable llm serving: A real-world workload study.
\newblock \emph{arXiv e-prints}, pages arXiv--2401.

\bibitem[{Weidinger et~al.(2022)Weidinger, Uesato, Rauh, Griffin, Huang, Mellor, Glaese, Cheng, Balle, Kasirzadeh et~al.}]{weidinger2022taxonomy}
Laura Weidinger, Jonathan Uesato, Maribeth Rauh, Conor Griffin, Po-Sen Huang, John Mellor, Amelia Glaese, Myra Cheng, Borja Balle, Atoosa Kasirzadeh, and 1 others. 2022.
\newblock Taxonomy of risks posed by language models.
\newblock In \emph{Proceedings of the 2022 ACM conference on fairness, accountability, and transparency}, pages 214--229.

\bibitem[{Wolf et~al.(2020)Wolf, Debut, Sanh, Chaumond, Delangue, Moi, Cistac, Rault, Louf, Funtowicz, Davison, Shleifer, von Platen, Ma, Jernite, Plu, Xu, Scao, Gugger, Drame, Lhoest, and Rush}]{wolf-etal-2020-transformers}
Thomas Wolf, Lysandre Debut, Victor Sanh, Julien Chaumond, Clement Delangue, Anthony Moi, Pierric Cistac, Tim Rault, Rémi Louf, Morgan Funtowicz, Joe Davison, Sam Shleifer, Patrick von Platen, Clara Ma, Yacine Jernite, Julien Plu, Canwen Xu, Teven~Le Scao, Sylvain Gugger, and 3 others. 2020.
\newblock \href {https://www.aclweb.org/anthology/2020.emnlp-demos.6} {Transformers: State-of-the-art natural language processing}.
\newblock In \emph{Proceedings of the 2020 Conference on Empirical Methods in Natural Language Processing: System Demonstrations}, pages 38--45, Online. Association for Computational Linguistics.

\bibitem[{Zhang et~al.(2024)Zhang, Wang, Li, Shou, Chen, Chen, and Mehrotra}]{zhang2024draft}
Jun Zhang, Jue Wang, Huan Li, Lidan Shou, Ke~Chen, Gang Chen, and Sharad Mehrotra. 2024.
\newblock Draft\& verify: Lossless large language model acceleration via self-speculative decoding.
\newblock In \emph{Proceedings of the 62nd Annual Meeting of the Association for Computational Linguistics (Volume 1: Long Papers)}, pages 11263--11282.

\end{thebibliography}

\newpage
\appendix

\section{Related Work}
\subsection{Test-Time Alignment}
Early alignment research focused on post‑training LLMs with techniques like RLHF~\citep{ouyang2022training}, but the high cost of policy optimization methods like PPO~\citep{schulman2017proximal} and GRPO~\citep{shao2024deepseekmath} has motivated the test‑time (or decoding‑time) alignment approaches that leave the LLMs' parameters frozen. For example, best-of-$N$ (BoN) generates multiple complete responses and select the highest‑reward one; rejection sampling continues generating proposal responses until a reward threshold is met. Built upon these 2 techniques, test-time alignment methods at varying 
granularities are developed. For example, token‑level reward‑guided search~\citep{deng2023reward, khanov2024args}, segment-level rejection sampling~\citep{li2024cascade}, and segment-level MC tree search~\citep{qiu2024treebon} have all been explored. However, although they improve the efficiency of test-time alignment, they still rely on external reward models to interact with the decoding process, which is the major cause of the inefficiency. Our algorithm (SSS) jumps out of the frameworks with reward models, and leverages a small draft model to reflect human preference. The interaction between draft and target models are far more efficient than the traditional LLM-RM framework.

\subsection{Speculative Sampling}
Orthogonal to test-time alignment, speculative decoding~\citep{chen2023accelerating, leviathan2023fast} accelerates LLM generation by using a lightweight draft model to guess $K$ future tokens, which is then verified by a large target model in parallel. Speculative sampling recovers the exact distribution of the target model~\citep{chen2023accelerating}. Recent work has aimed to further improve the efficiency of this draft–verify framework. One major direction is increasing the number of candidate tokens accepted by the target model. To this end, tree-based methods~\citep{miao2024specinfer, li2024eagle, fu2024lookahead} generate multiple draft token paths in parallel, increasing the likelihood of acceptance and reducing verification overhead. Other studies enhance draft token quality through knowledge distillation~\citep{liu2024online}, layer-skipped decoding~\citep{elhoushi2024layerskip, zhang2024draft}, or by adding specialized speculative heads such as MEDUSA~\citep{cai2024medusa} to improve token prediction and verification efficiency. Some recent variants also explore using reward models (RMs) as verifiers instead of target LLMs~\citep{liao2025reward}. This change allows for higher token acceptance rates by relaxing fluency constraints. \citet{nakshatri2024constrained} applies speculative sampling to accelerate constrained decoding, still relying on external reward models to enforce the constraints. Despite these advances, existing speculative decoding methods are primarily designed for computational acceleration and assume an \emph{unaligned} draft model. As a result, the question of how to incorporate human preference alignment into the speculative decoding process without sacrificing efficiency remains largely under-explored.

\section{Proofs}
\subsection{RLHF Optimal Solution\label{sec:proof_rlhf}}
Starting from the RLHF objective as demonstrated in Eq.~\eqref{eq:rlhf}, the KL-constrained reward maximization can be re-written as:
\begin{equation}
\begin{aligned}
    &\frac{1}{\beta}\mathbb{E}_{y\sim \pi_{\param}(\cdot|x)}r(x,y) - \mathbf{KL}(\pi_{\param}(\cdot|x) \| \pi_{\text{ref}}(\cdot|x)) \\
    &= \sum_{y}\pi_{\param}(y|x)\cdot\left(\log\frac{\pi_{\text{ref}}(y|x)}{\pi_{\param}(y|x)} + \frac{1}{\beta}r(x,y)\right) \\
    &= \sum_{y}\pi_{\param}(y|x)\cdot\log\frac{\pi_{\text{ref}}(y|x)\exp\left(\frac{1}{\beta}r(x,y)\right)}{\pi_{\param}(y|x)} \\
    &\propto -\mathbf{KL}(\pi_{\param}(\cdot|x) \| \pi_{\text{ref}}(y|x)\exp\left(\frac{1}{\beta}r(x,y)\right)).
\end{aligned}
\end{equation}
Optimizing a model $\pi_{\param}$ with RLHF is equivalent to minimizing its KL-divergence from a new policy: $\pi^\star(y|x) = \pi_{\text{ref}}(y|x)\exp\left(\frac{1}{\beta}r(x,y)\right)$, which we call the RLHF optimal solution.

\subsection{Theorem~\ref{theorem}\label{sec:proof_theorem}}
The following proof is inspired by Theorem 1 of \citet{chen2023accelerating}. We start by considering the probability of generating a token $x$, and ignore the prompt and previously generated tokens for simplicity. In our shifted speculative sampling as demonstrated in Algorithm~\ref{algo:sss}, a token can be generated in two cases: \romannumeral1) it is a draft token and is accepted, and \romannumeral1) draft token is rejected and it is sampled as a bonus token. We show these two case in the following formula:
\begin{equation}
\begin{aligned}
    &\mathbf{P}(X=x) \\
    &= \pi_{\text{draft}}^r(\hat{x}=x) \cdot \mathbf{P}(\hat{x}~\text{accepted}|\hat{x}=x) \\
    &~~~~+ \mathbf{P}(\hat{x}~\text{rejected}) \cdot \mathbf{P}(X=x|\hat{x}~\text{rejected}).
\end{aligned}
\end{equation}
The left half can be re-written as:
\begin{equation}
\begin{aligned}
    &\pi_{\text{draft}}^r(\hat{x}=x) \cdot \mathbf{P}(\hat{x}~\text{accepted}|\hat{x}=x) \\
    &\overset{\text{(a)}}{=}\pi_{\text{draft}}^r(\hat{x}=x) \cdot \min\left(1, \frac{\pi_{\text{ref}}(x)}{\pi_{\text{draft}}^{\text{SFT}}(x)}\right) \\
    &=\pi_{\text{draft}}^r(\hat{x}=x) \cdot \min\left(1, \frac{\pi^\star(x)}{\pi_{\text{draft}}^r(x)}\right) \\
    &=\min\left(\pi_{\text{draft}}^r(x), \pi^\star(x)\right).
\end{aligned}
\end{equation}
Here, (a) is from Eq.~\eqref{eq:accept_prob_new}, where we design the specific form for the proof completeness. For the right half, we first transform the probability of rejection to be:
\begin{equation}
\begin{aligned}
    \mathbf{P}(\hat{x}~\text{rejected}) &= 1 - \mathbf{P}(\hat{x}~\text{accepted}) \\
    &= 1 - \sum_{x'}\mathbf{P}(\hat{x}=x', \hat{x}~\text{accepted}) \\
    &= 1 - \sum_{x'}\min\left(\pi_{\text{draft}}^r(x'), \pi^\star(x')\right) \\
    &= \sum_{x'} \pi^\star(x') - \min\left(\pi_{\text{draft}}^r(x'), \pi^\star(x')\right) \\
    &=\sum_{x'}\max\left(0, \pi^\star(x') - \pi_{\text{draft}}^r(x')\right).
\end{aligned}
\end{equation}
Then, the probability of generating $x$ after draft token rejection is:
\begin{equation}
\begin{aligned}
    \mathbf{P}(X=x|\hat{x}~\text{rejected}) &= \pi_{\text{bonus}}^r(x) \\
    &= \left(\pi_{\text{draft}}^r(x) \left(\frac{\pi_{\text{ref}}(x)}{\pi_{\text{draft}}^{\text{SFT}}(x)} - 1\right)\right)_+ \\
    &= (\pi^\star(x) - \pi_{\text{draft}}^r(x))_+ \\
    &= \frac{\max\left(0, \pi^\star(x) - \pi_{\text{draft}}^r(x)\right)}{\sum_{x'}\max\left(0, \pi^\star(x') - \pi_{\text{draft}}^r(x')\right)},
\end{aligned}
\end{equation}
where $(f(x))_+=\max(0,f(x))/\sum_x\max(0,f(x))$ is the clamp normalization operator. Finally, taking all above together, the token generation probability is:
\begin{equation}
\begin{aligned}
    &\mathbf{P}(X=x) \\
    &= \min\left(\pi_{\text{draft}}^r(x), \pi^\star(x)\right) + \max\left(0, \pi^\star(x) - \pi_{\text{draft}}^r(x)\right) \\
    &= \pi^\star(x),
\end{aligned}
\end{equation}
always equivalent to the RLHF optimal solution.

\section{Why SSS Achieves Alignment Quality and Inference Efficiency Simultaneously?}
As shown in \Cref{tab:reward_comparison}, our proposed method (SSS) consistently outperforms existing test-time alignment approaches in both alignment quality and decoding efficiency. This is achieved through two key design choices:
\begin{itemize}
    \item \textbf{Draft model alignment reduces reward evaluation cost.} By aligning a small draft model to human preference using DPO~\citep{rafailov2023direct}, SSS avoids frequent reward model queries during decoding, significantly lowering computational overhead compared to segment-level or prefix-level alignment methods like CARDS~\citep{li2024cascade} and TreeBoN~\citep{qiu2024treebon}.
    \item \textbf{Speculative sampling is adapted for alignment.} Standard speculative decoding accelerates generation but lacks alignment. SSS modifies the acceptance rule and residual distribution to account for the distributional shift caused by draft model alignment, enabling efficient decoding while preserving preference consistency.
\end{itemize}
As a result, SSS achieves fast and well-aligned generation without requiring an aligned target model, and effectively recovers the RLHF optimal solution.

\section{Ablation Studies\label{sec:exp_ablation}}
\subsection{How to select a good draft model?}
\Cref{tab:draft_selection_compact} presents the ablation studies comparing the key indicators of draft model training across three models. We evaluate each draft model using chosen/rejected/generated likelihoods, implicit reward accuracy, and gold reward. Pretrained draft models tend to achieve higher implicit reward accuracy, especially on \texttt{OPT-125M} and \texttt{OPT-350M}, but often produce lower gold rewards, indicating misalignment with human preferences despite higher token-level likelihood. SFT models sometimes yield stronger gold rewards (e.g., on \texttt{OPT-350M}), but the results are not always consistent across models. DPO consistently delivers the best trade-off across all checkpoints. This highlights that no single metric fully captures alignment quality, and selecting a good draft model requires balancing alignment quality, generation fluency, and reward supervision.

\begin{table}[ht]\scriptsize\setlength{\tabcolsep}{4pt}
\centering
\caption{
\textbf{Performance of different draft model training methods using HH-RLHF.}
"Lik(Chosen)", "Lik(Rej)", and "Lik(Gen)" denote the average likelihood of the chosen, rejected, and generated responses, respectively. 
"Imp. Rw" refers to implicit reward accuracy, and "Gold Rw" denotes the average reward from a gold reward model.
}
\begin{tabular}{c|ccccc}
\toprule
\textbf{Method} & \textbf{Lik(Chosen)} & \textbf{Lik(Rej)} & \textbf{Lik(Gen)} & \textbf{Imp. Rw} & \textbf{Gold Rw} \\
\midrule
\multicolumn{6}{c}{\texttt{Qwama-0.5B}} \\
\midrule
DPO        & 0.5428 & 0.4984 & 0.6809 & 0.43 & 3.75 \\
SFT        & 0.4973 & 0.4607 & 0.6314 & 0.43 & 3.49 \\
Pretrained & 0.3448 & 0.3314 & 0.5402 & 0.42 & 4.29 \\
\midrule
\multicolumn{6}{c}{\texttt{OPT-125M}} \\
\midrule
DPO        & 0.3944 & 0.3999 & 0.6285 & 0.54 & 3.18 \\
SFT        & 0.3542 & 0.3546 & 0.6282 & 0.57 & 3.55 \\
Pretrained & 0.3071 & 0.3003 & 0.6156 & 0.59 & 3.46 \\
\midrule
\multicolumn{6}{c}{\texttt{OPT-350M}} \\
\midrule
DPO        & 0.3251 & 0.3139 & 0.5622 & 0.62 & 3.45 \\
SFT        & 0.3875 & 0.3881 & 0.6343 & 0.55 & 3.59 \\
Pretrained & 0.3307 & 0.3226 & 0.6174 & 0.58 & 3.35 \\
\bottomrule
\end{tabular}
\label{tab:draft_selection_compact}
\end{table}

\subsection{Practical tricks to handle the gap between DPO draft models and the desired well-aligned draft model}
\begin{figure}[ht]
    \centering
    \includegraphics[width=\linewidth]{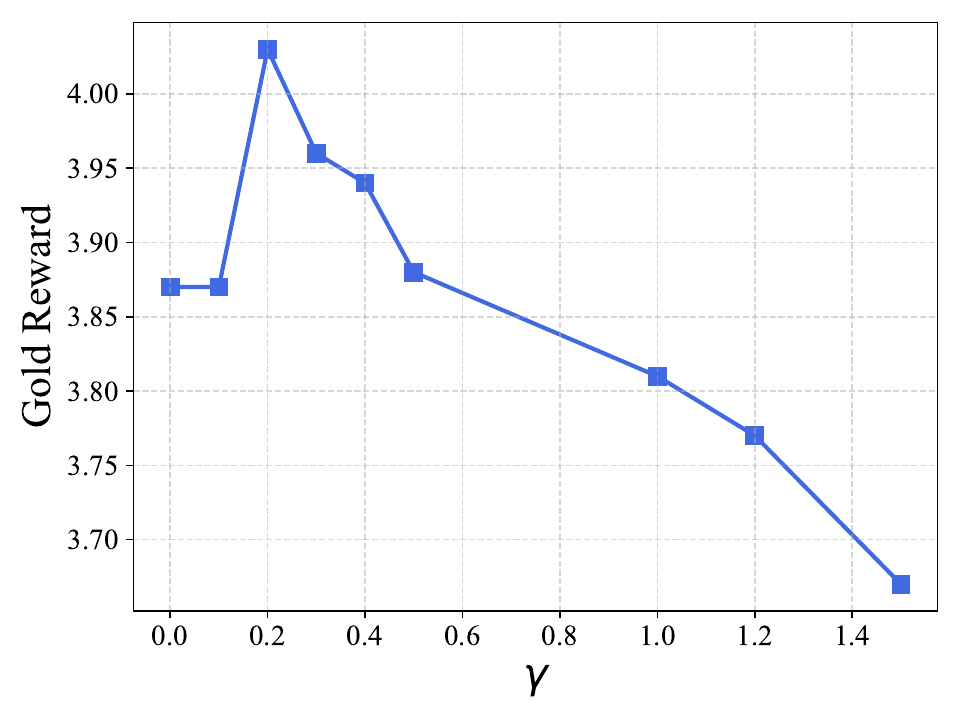}
    \caption{Effect of hyper-parameter $\gamma$ to the gold reward on \texttt{OPT-13B/OPT-350M} model. The optimal $\gamma$ is below 0.5.}
    \label{fig:gamma}
\end{figure}

The proposed algorithm relies on the assumption that we have a well-aligned draft model $\pi_{\text{draft}}^r(y|x) \approx \pi_{\text{draft}}^{\text{SFT}}(y|x)\cdot\exp\left(\frac{1}{\beta}r(x,y)\right)$. However, this is often hard to obtain and verify since we do not have access to the reward score $r(x,y)$. We find that, even for imperfect shifted reward models, we can still have outstanding alignment quality by slightly modifying Eq.~\eqref{eq:r_bonus} to be:
\begin{equation}
\begin{aligned}
    \left(\pi_{\text{draft}}^r(\cdot|x,y_{<t+K'})^\gamma \left(\frac{\pi_{\text{ref}}(\cdot|x,y_{<t+K'})}{\pi_{\text{draft}}^{\text{SFT}}(\cdot|x,y_{<t+K'})} - 1\right)\right)_+,
\end{aligned}
\end{equation}
where $\gamma=1$ is exactly the original version. We test a set of $\gamma$ values and find that the optimal $\gamma$ is below 0.5, as shown in Fig.~\ref{fig:gamma}.

\subsection{Is SSS Really Approaching the RLHF Optimal Solution?}
As this paper claims that SSS can recover the RLHF optimal solution, we compare two RLHF-trained target models (DPO\footnote{\url{https://huggingface.co/Nagi-ovo/Llama-3-8B-DPO}.} and PPO\footnote{\url{https://huggingface.co/OpenRLHF/Llama-3-8b-rlhf-100k}.}) with SSS in Table~\ref{tab:rlhf_optimal}. The results demonstrate that SSS achieves a better gold reward than the trained target models. It is noteworthy that the RLHF-trained target models are all approximations to the optimal solution, and therefore are they presenting lower gold rewards than SSS.
\begin{table}
\caption{Comparison between RLHF-trained target models and SSS. The proposed SSS achieves better gold reward than RLHF-trained target models.}
\label{tab:rlhf_optimal}
\centering
\begin{tabular}{lc}
\hline
\multicolumn{1}{c}{\textbf{Method}} & \textbf{Gold Reward} \\ \hline
DPO                                 & 4.85                 \\
PPO                                 & 5.62                 \\
SSS (ours)                          & 6.14                 \\ \hline
\end{tabular}
\end{table}

\begin{figure*}
    \centering
    \includegraphics[width=\linewidth]{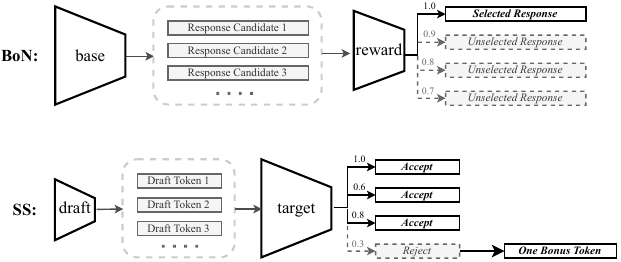}
    \caption{Inference procedures of best-of-$N$ (traditional test-time alignment) and speculative sampling (our new algorithm). Speculative sampling leverages a small draft model to efficiently guess future tokens, which has the potential to be shifted to human preference.}
    \label{fig:bon_ss}
\end{figure*}

\begin{algorithm}
\caption{Speculative Sampling}
\label{algo:ss}
\begin{algorithmic}

\Require autoregressive target model $\pi_{\text{ref}}$ and autoregressive draft model $\pi_{\text{draft}}$;
\Require initial prompt $x$ and lookahead $K>0$.

\State $y \gets \emptyset$
\State $t \gets 0$

\While{$t < \texttt{max\_length}$}
    \For{k = 1 : K}
        \State $\hat{y}_{t+k} \sim \pi_{\text{draft}}(\cdot|x,y_{<t},\hat{y}_{t:t+k-1})$
    \EndFor
    \State $\pi_{\text{ref}}(\hat{y}_{t:t+K}|x,y_{<t})$ \Comment{Target Likelihoods}

    \For{k = 1 : K}
        \State $\epsilon \sim \mathcal{U}[0,1]$
        \If{$\epsilon < p_{\text{accept}}(t)$ in Eq.~\eqref{eq:accept_prob}}
            \State $y_{t+k} \gets \hat{y}_{t+k}$ \Comment{Accept}
        \Else
            \State $y_{t+k} \gets \pi_{\text{bonus}}(\cdot|x,y_{<t+k})$ in Eq.~\eqref{eq:bonus}
            \State $t \gets t+k$ and break
        \EndIf
    \EndFor

    \If{all draft tokens are accepted}
        \State $y_{t+K+1} \sim \pi_{\text{ref}}(\cdot|x,y_{\leq t+K})$
        \State $t \gets t+K+1$
    \EndIf
\EndWhile

\end{algorithmic}
\end{algorithm}

\end{document}